\pdfoutput=1
\PassOptionsToPackage{table, dvipsnames}{xcolor}

\documentclass[11pt]{article}

\usepackage{xcolor}
\usepackage[final]{acl}

\usepackage{times}
\usepackage{latexsym}
\usepackage{amsfonts}

\usepackage[T1]{fontenc}

\usepackage[utf8]{inputenc}

\usepackage{microtype}

\usepackage{inconsolata}

\usepackage{graphicx}
\usepackage{amsmath}
\usepackage{enumitem}
\usepackage{booktabs}
\usepackage{adjustbox}
\usepackage{multirow}
\usepackage{makecell}
\usepackage{bbding}
\newcommand{\Sref}[1]{\S\ref{#1}}

\title{Can LLM Graph Reasoning Generalize beyond Pattern Memorization?}

\author{Yizhuo Zhang\thanks{equal contribution}\textsuperscript{1} \ \ \ \ \ \ \ \ Heng Wang\footnotemark[1]\textsuperscript{2} \ \ \ \ \ \ \ \ Shangbin Feng\footnotemark[1]\textsuperscript{1} \\
\textbf{Zhaoxuan Tan\textsuperscript{3}} \ \ \ \textbf{Xiaochuang Han\textsuperscript{1}} \ \ \ \textbf{Tianxing He\textsuperscript{4}} \ \ \ \textbf{Yulia Tsvetkov\textsuperscript{1}} \\
\textsuperscript{1}University of Washington \ \ \textsuperscript{2}Xi'an Jiaotong University \\
\textsuperscript{3}University of Notre Dame \ \ \textsuperscript{4}Tsinghua University \\
\href{mailto:mattyz@uw.edu}{\texttt{mattyz@uw.edu}} \ \ \href{mailto:wh2213210554@stu.xjtu.edu.cn}{\texttt{wh2213210554@stu.xjtu.edu.cn}} \ \ \href{mailto:shangbin@cs.washington.edu}{\texttt{shangbin@cs.washington.edu}}
}

\newcommand{\ourmethod}[1]{\textsc{NLGift}}
\newcommand{\lowbar}[1]{\textit{Significant Transfer}}
\newcommand{\highbar}[1]{\textit{Strong Recovery}}
\definecolor{lowbar_color}{rgb}{0.851,0.918,0.827}
\definecolor{highbar_color}{rgb}{0.576,0.769,0.49}
\definecolor{diag_color}{rgb}{0.718,0.718,0.718}

\begin{document}
\maketitle

\begin{abstract}
Large language models (LLMs) demonstrate great potential for problems with implicit graphical structures, while recent works seek to enhance the graph reasoning capabilities of LLMs through specialized instruction tuning. The resulting ``graph LLMs'' are evaluated with in-distribution settings only, thus it remains underexplored whether LLMs are learning generalizable graph reasoning skills or merely memorizing patterns in the synthetic training data. To this end, we propose the \ourmethod{} benchmark, an evaluation suite of LLM graph reasoning generalization: whether LLMs could go beyond \emph{semantic}, \emph{numeric}, \emph{structural}, \emph{reasoning} patterns in the synthetic training data and improve utility on \emph{real-world} graph-based tasks. Extensive experiments with two LLMs across four graph reasoning tasks demonstrate that while generalization on simple patterns (semantic, numeric) is somewhat satisfactory, LLMs struggle to generalize across reasoning and real-world patterns, casting doubt on the benefit of synthetic graph tuning for real-world tasks with underlying network structures. We explore three strategies to improve LLM graph reasoning generalization, and we find that while post-training alignment is most promising for real-world tasks, empowering LLM graph reasoning to go beyond pattern memorization remains an open research question.\footnote{Code and data are publicly available at \href{https://github.com/MatthewYZhang/NLGift}{https://github.com/MatthewYZhang/NLGift}.}
\end{abstract}

\section{Introduction}
Large Language Models (LLMs) are increasingly employed for tasks at the intersection of language and structure such as multi-hop QA \citep{strategyqa, ding2023knowledge} and structured commonsense reasoning \citep{sakaguchi2021proscript, madaan-etal-2022-language}. These problems are often described in natural language and have implicit graphical structures \citep{strategyqa, ding2023knowledge, saha2021explagraphs, sakaguchi2021proscript}, where LLMs' graph reasoning capabilities are tested. While LLMs \emph{do} possess preliminary abilities to represent and reason with graphs \citep{wang2023can}, they also face challenges such as hallucinations \citep{huang2023survey} and prompt sensitivity \citep{fatemi2024talk} when dealing with structured data.

\begin{figure*}[h]
    \centering
    \includegraphics[width=1\linewidth]{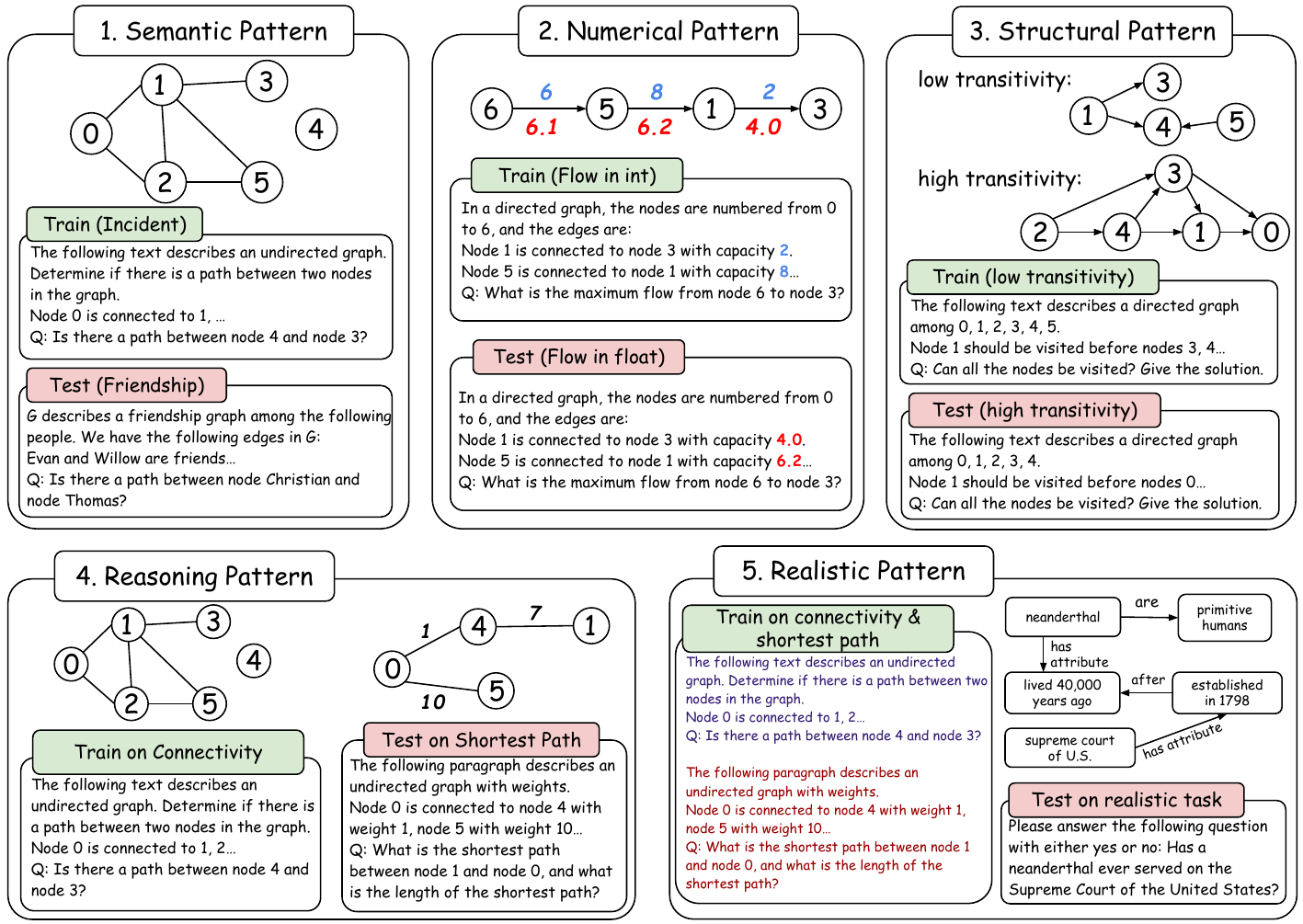}

    \caption{Overview of the \ourmethod{} Benchmark, featuring five types of graph reasoning patterns that are increasingly challenging in order. We present an example for each pattern to show the transfer from training to test sets.}
    \label{fig:overview}
\end{figure*}

Existing works seek to improve LLM graph reasoning mainly through better prompting \citep{fatemi2024talk} or instruction tuning \citep{wang2024instructgraph, chen2024graphwiz}, while the latter line of training-based approaches is generally more effective in producing specialized models for graph-based applications \citep{tang2023graphgpt, wang2024instructgraph, chen2024graphwiz, luo2024graphinstruct}. However, these approaches are often evaluated in in-distribution settings, while robust graph reasoners should go beyond training sets to encode general and transferable graph reasoning capabilities in model parameters. Consequently, we ask: \emph{Are LLMs graph reasoners or merely pattern regurgitators?} More concretely, \emph{Can LLM graph reasoning go beyond memorizing patterns in the training data and perform well in out-of-distribution contexts?} The answer to this question has profound implications for LLM reliability in structured contexts since real-world graph-based problems are diverse, heterogeneous, and constantly evolving.

To this end, we propose \textbf{\ourmethod{}}, a comprehensive testbed of \textbf{N}atural \textbf{L}anguage \textbf{G}raph reasoning with sh\textbf{ift}ing patterns. \ourmethod{} contains 37,000 problems in total, where LLMs are instruction tuned on a subset of problems with distribution \( \mathcal{D}_{train} \) and evaluated on both in-distribution \( \mathcal{D}_{train} \) and out-of-distribution \( \mathcal{D}_{test} \) test sets. \ourmethod{} features five types of patterns where LLMs should generalize beyond: 1) \emph{semantic}, 2) \emph{numerical}, 3) \emph{structural}, 4) \emph{reasoning}, and 5) \emph{real-world patterns}. The first four settings focus on transferring across patterns in the synthetic graph data, while the real-world pattern focuses on the transfer from synthetic graph problems to real-world tasks with graph implications such as structured commonsense reasoning \citep{saha2021explagraphs, sakaguchi2021proscript} and multi-hop QA \citep{strategyqa, ding2023knowledge}. These five patterns are increasingly challenging in order and offer a progressive testbed of LLM graph reasoning generalization.

To quantify the success of generalization, \ourmethod{} establishes two standards: 1) the \emph{basic} standard: \lowbar{}, i.e., the tuned model's improvement on out-of-distribution test sets over the zero-shot untuned model is statistically significant; 2) the \emph{strong} standard: \highbar{}, i.e., when evaluated on out-of-distribution data, the tuned LLMs could substantially recover the gains of in-distribution training. Robust graph reasoners should ideally meet both \emph{basic} and \emph{strong} generalization standards across the five patterns.

Extensive experiments on \ourmethod{} with two LLMs demonstrate that on easier patterns (semantic, numerical, and structural), LLMs achieve the \emph{basic} standard 75\% of the time but only reach the \emph{strong} standard in 35\% of settings. On reasoning patterns, LLMs only achieve \emph{basic} generalization 33\% of the time while \emph{never} qualify for the \emph{strong} standard of generalization. What's worse, on the most challenging real-world patterns, LLMs achieve \emph{basic} generalization in 6\% of the settings while graph instruction tuning is counterproductive in 69\% of cases, casting doubt on the benefit of synthetic graph data. Further analysis reveals that task composition and keyword frequency in training corpus greatly impact graph reasoning generalization.

We explore three preliminary strategies to augment LLM graph reasoning generalization: \emph{code mixing}, \emph{machine-generated CoTs}, and \emph{post-training alignment}. While post-training alignment is most promising on real-world tasks, empowering LLMs with general and transferable graph reasoning abilities remains an open research question. To sum up, our key contributions include presenting the \ourmethod{} benchmark, evaluating LLM graph reasoning generalization with diverse tasks and patterns, as well as exploring preliminary solutions to mitigate the profound generalization challenges.

\section{\ourmethod{} Benchmark}
\label{sec:dataset}



To examine whether LLMs are capable of robust reasoning with graph problems rather than memorizing training patterns, we present the \ourmethod{} Benchmark (Figure \ref{fig:overview}). Specifically, we select four representative graph reasoning tasks: connectivity, shortest path, topological sort, and maximum flow. We then design five patterns that LLMs should generalize beyond. We present the five patterns in ascending order of difficulty so that they can serve as a progressive testbed.

\subsection{Graph reasoning patterns}
\label{sec:reasoning pattern}

\paragraph{Semantic Patterns}

The semantic pattern involves representing the graph problem with different natural language descriptions such as ``edge (2,3)'' or ``Bob and Amy are friends''.
When trained on one type of semantic representation, robust graph reasoners should achieve similar levels of performance when testing with different semantic representations.
Specifically, we employ four typical representation methods: adjacency, incident, graph expert, and friendship \citep{fatemi2024talk}. 

\begin{itemize}[leftmargin=*]
\setlength\itemsep{-0.3em}
    \item \emph{Adjacency}: We list all the edges in the graph with natural language. (i.e., node 1 is connected to node 3. node 1 is connected to 2)
    \item \emph{Incident}: We describe the connectivity of each node in a single sentence. (i.e., node 1 is connected to 2, 3, 4)
    \item \emph{Graph Expert}: We employ the prompt ``You are a graph analyst'' as a prefix, and then use letters to represent the nodes. (i.e., A->B, A->C)
    \item \emph{Friendship}: We describe the nodes as people and the edges as the friendship between people. (i.e., Alice and Carol are friends)
\end{itemize}

\paragraph{Numerical Patterns}

For the numerical pattern, we aim to investigate whether different number distributions in the edge attributes (i.e., edge weight, edge capacity) might affect LLM graph reasoning. Specifically, we employ three different number distributions: small integers (from 1 to 10), large integers (from 11 to 100), and floats with one floating point (from 1.0 to 10.0). Trained on graph problems with one numerical distribution of node/edge attributes, robust graph reasoners should achieve similar performance on another set of problems with different numerical distributions.

\paragraph{Structural Patterns}

Graphs are often diverse and heterogeneous, featuring varying levels of size, centrality, and other structural variations. We therefore investigate the impact of structural patterns, if changes in graph structural properties might affect the performance of LLM's reasoning abilities. We specifically design three different criteria to quantify a network's structural features. 

\begin{itemize}[leftmargin=*]
    \item \emph{Graph Size}: Small-size graphs have 3 to 10 nodes while large-size graphs have 11 to 25 nodes.
    \item \emph{Graph Generator}: we use two different graph generator algorithms to generate graphs, specifically, Erdos-Renyi \citep{erdHos1960evolution} and Barabasi-Albert \citep{barabasi1999emergence}.
    \item \emph{Graph Transitivity}: we calculate the graph transitivity \cite{luce1949method, wasserman1994social} and partition them into low-transitivity and high-transitivity subsets. Graph transitivity (T) is calculated as $\mathrm{T} = \frac{3 \times \text{number of triangles}}{\text{number of all triplets}}$.
\end{itemize}

\paragraph{Reasoning Patterns}

For the previous three patterns, we train and test LLMs on the same graph tasks, while a robust graph reasoner should learn universal principles of graph reasoning to generalize across different graph reasoning problems. To evaluate this, we use the four graph reasoning tasks (\Sref{sec:dataset}), instruction-tune the LLM on one synthetic graph task and evaluate it on both the same task and three other graph tasks.

\begin{table*}[h]
  \centering
  
  \begin{adjustbox}{max width=1\textwidth}
    \begin{tabular}{c c c c c c c c c}
    \toprule[1.5pt]
    \multirow{2}{*}{train/test} & \multicolumn{4}{c}{Connectivity} & \multicolumn{4}{c}{Shortest Path} \\
    \cmidrule(lr){2-5} \cmidrule(lr){6-9}
    & Adjacency & Friendship & Expert & Incident & Adjacency & Friendship & Expert & Incident \\
    \midrule
    \multicolumn{9}{c}{\textbf{\textsc{LLaMA2-7B}}} \\

    \midrule[0.2pt]
    \textsc{Adjacency} & \cellcolor{diag_color}.672 (+29\%) & .518 (0\%) & .512 (-2\%) & \cellcolor{lowbar_color}.660 (+27\%) & \cellcolor{diag_color}.212 (+1225\%) & \cellcolor{lowbar_color}.122 (+578\%) & \cellcolor{highbar_color}.152 (+850\%) & \cellcolor{lowbar_color}.132 (+725\%) \\
    \textsc{Friendship} & .500 (-4\%) & \cellcolor{diag_color}.694 (+34\%) & \cellcolor{highbar_color}.686 (+31\%) & .540 (+4\%) & .028 (+75\%) & \cellcolor{diag_color}.242 (+1244\%) & \cellcolor{lowbar_color}.086 (+438\%) & .042 (+163\%) \\
    \textsc{Expert} & .482 (-8\%) & .548 (+6\%) & \cellcolor{diag_color}.624 (+20\%) & .484 (-7\%) & \cellcolor{lowbar_color}.044 (+175\%) & .028 (+56\%) & \cellcolor{diag_color}.184 (+1050\%) & \cellcolor{lowbar_color}.046 (+188\%) \\
    \textsc{Incident} & \cellcolor{highbar_color}.662 (+27\%) & .512 (-1\%) & .502 (-4\%) & \cellcolor{diag_color}.728 (+40\%) & \cellcolor{lowbar_color}.132 (+725\%) & \cellcolor{lowbar_color}.084 (+367\%) & \cellcolor{lowbar_color}.124 (+675\%) & \cellcolor{diag_color}.244 (+1425\%) \\
    \midrule[0.2pt]
    \textsc{Zero-shot} & .522 & .516 & .522 & .520 & .016 & .018 & .016 & .016 \\
    \midrule[1.2pt]

    \multicolumn{9}{c}{\textbf{\textsc{ChatGPT}}} \\
    \midrule[0.2pt]
    \textsc{Adjacency} & \cellcolor{diag_color}.958 (+25\%) & \cellcolor{lowbar_color}.826 (+12\%) & \cellcolor{lowbar_color}.880 (+16\%) & \cellcolor{lowbar_color}.912 (+24\%) & \cellcolor{diag_color}.542 (+132\%) & \cellcolor{highbar_color}.516 (+153\%) & \cellcolor{highbar_color}.512 (+129\%) & \cellcolor{highbar_color}.546 (+101\%) \\
    \textsc{Friendship} & \cellcolor{lowbar_color}.892 (+17\%) & \cellcolor{diag_color}.950 (+29\%) & \cellcolor{highbar_color}.892 (+18\%) & \cellcolor{lowbar_color}.896 (+21\%) & \cellcolor{lowbar_color}.426 (+82\%) & \cellcolor{diag_color}.476 (+133\%) & \cellcolor{highbar_color}.486 (+117\%) & \cellcolor{lowbar_color}.518 (+90\%) \\
    \textsc{Expert} & \cellcolor{highbar_color}.930 (+22\%) & \cellcolor{lowbar_color}.874 (+19\%) & \cellcolor{diag_color}.922 (+22\%) & \cellcolor{lowbar_color}.916 (+24\%) & \cellcolor{lowbar_color}.472 (+102\%) & \cellcolor{lowbar_color}.400 (+96\%) & \cellcolor{diag_color}.526 (+135\%) & \cellcolor{highbar_color}.568 (+109\%) \\
    \textsc{Incident} & .808 (+6\%) & .780 (+6\%) & .772 (+2\%) & \cellcolor{diag_color}.964 (+31\%) & \cellcolor{lowbar_color}.324 (+38\%) & \cellcolor{lowbar_color}.390 (+91\%) & \cellcolor{lowbar_color}.354 (+58\%) & \cellcolor{diag_color}.616 (+126\%) \\
    \midrule[0.2pt]
    \textsc{Zero-shot} & .764 & .736 & .756 & .738 & .234 & .204 & .224 & .272 \\
    \bottomrule[1.5pt]
    \end{tabular}%
    \end{adjustbox}
    \caption{Results for the semantic pattern where colors indicate \colorbox{lowbar_color}{\lowbar{}},\colorbox{highbar_color}{\highbar{}}, and \colorbox{diag_color}{in-distribution results}. LLMs can generalize across different semantic representations of the same graph problems to some extent and achieve \lowbar{} 68.8\% of the time. The larger \textsc{ChatGPT} is stronger in generalization, achieving \lowbar{} and \highbar{} in 21 and 7 cases compared to \textsc{LLaMA2}'s 12 and 3.}
  \label{tab:semantic-result}
\end{table*}

\paragraph{Real-world Patterns}

While the previous four patterns all train and test on synthetic graph data, we argue that \emph{the ultimate goal of graph synthetic tuning is to benefit real-world problems with underlying graph structures}: after all, these synthetic problems could be solved with 100\% accuracy by conventional algorithms.
Thus for the real-world pattern, we fine-tune LLMs with synthetic graph data, and then evaluate with real-world problems that have implicit graph structures. We specifically employ two types of datasets for evaluation:

\begin{itemize}[leftmargin=*]
\setlength\itemsep{-0.1em}
    \item \emph{Multi-Hop QA}: Multi-hop QA involves answering questions that require multi-hop reasoning, which is inherently related to synthetic graph problems such as connectivity or shortest path, as the solving process can be viewed as navigating through a network of concepts and relations or trying to find the shortest path between two concepts using existing relations. We adopt StrategyQA \citep{strategyqa} and Knowledge Crosswords \citep{ding2023knowledge} for the multi-hop QA task. 

    \item \emph{Structured Commonsense Reasoning}: This task tests the intersection of commonsense knowledge and structured reasoning, where LLMs need to figure out the dependency structure of events (\emph{e.g.} open the fridge before taking milk from the fridge) and generate a reasonable plan. This is strongly correlated with synthetic graph problems such as topological sort, where different steps have a constraint of order. We specifically employ ExplaGraphs \citep{saha2021explagraphs} and Proscript \citep{sakaguchi2021proscript} for this task.
\end{itemize}
Finally, we obtain the \ourmethod{} Benchmark with 33,000 synthetic problems and 4,000 realistic problems. Details of \ourmethod{} and the four real-world datasets can be found in Appendix \ref{sec:real-world dataset}.


\subsection{Generalization Metrics}
We instruction-tune a pre-trained language model \( f_{\theta} \) with initial parameters \( \theta \) on one distribution of training data \( \mathcal{D}_{train} \). 
Formally, this is expressed as:
\begin{align*}
    \theta_{\mathcal{D}_{train}} = \arg\min_{\theta} \mathbb{E}_{(x,y) \sim \mathcal{D}_{train}} [L(f_{\theta}(x), y)], 
\end{align*}
where \( L \) represents the loss function, and \( (x,y) \) are input-output pairs in instruction tuning.
We evaluate the tuned model on a different data distribution \( \mathcal{D}_{test} \) to test its generalization ability, which is defined as a \textit{generalization pair}.
\begin{align*}
    \mathrm{Acc}(\mathcal{D}_{\textit{test}} \mid \theta_{\mathcal{D}_{\textit{train}}}) = \mathbb{E}_{(x,y) \sim \mathcal{D}_{test}} \mathcal{I}(f_{\theta_{\mathcal{D}_{train}}}(x), y),
\end{align*}
where $\mathcal{I}(f_{\theta_{\mathcal{D}_{train}}}(x), y)$ is the indicator function of whether the $f_{\theta_{\mathcal{D}_{train}}}(x)$ and $y$ have the same reasoning path and reach the same answer.

We use two standards to quantify the success of LLM graph reasoning generalization: 1) \lowbar{} and 2) \highbar{}. Concretely, \lowbar{} is when the performance of the tuned model on $\mathcal{D}_{\textit{test}}$ is significantly better than the original untuned model, which is tested by proportional z-test \cite{statsmodels}:
\begin{align*}
    \mathrm{Acc}(\mathcal{D}_{\textit{test}} \mid \theta_{\mathcal{D}_{\textit{train}}}) \gg \mathrm{Acc}(\mathcal{D}_{\textit{test}} \mid \theta) \quad (\text{p} < 0.01)
\end{align*}
where $\mathrm{Acc}(\mathcal{D}_{\textit{test}} \mid \theta)$ refers to untuned model's performance on $\mathcal{D}_{test}$ data.

The higher criteria, \highbar{}, means the model's performance on out-of-distribution data should largely match the performance with in-distribution training. Specifically, we define Performance Gap Recovered (PGR) following \citet{burns2023weak}, and achieving \highbar{} requires that PGR is no less than a threshold parameter $\lambda$:
\begin{align*}
\mathrm{PGR} = \frac{\mathrm{Acc}(\mathcal{D}_{\textit{test}} \mid \theta_{\mathcal{D}_{\textit{train}}}) - \mathrm{Acc}(\mathcal{D}_{\textit{test}} \mid \theta)}{\mathrm{Acc}(\mathcal{D}_{\textit{test}} \mid \theta_{\mathcal{D}_{\textit{test}}}) - \mathrm{Acc}(\mathcal{D}_{\textit{test}} \mid \theta)}
\geq \lambda
\end{align*}
In the following experiments, we empirically set the PGR threshold $\lambda=0.8$ while we also experiment with other values in Appendix \ref{app:pgr-lambda-choice}.

\begin{table*}[t]
  \centering
  \begin{adjustbox}{max width=0.9\textwidth}
    \begin{tabular}{c c c c c c c}
    \toprule[1.5pt]
    \multirow{2}{*}{train/test} & \multicolumn{3}{c}{Shortest Path} & \multicolumn{3}{c}{Maximum Flow} \\
    \cmidrule(lr){2-4} \cmidrule(lr){5-7}
    
     & Small Int & Large Int & Float & Small Int & Large Int & Float \\
    
    \midrule
    \multicolumn{7}{c}{\textbf{\textsc{LLaMA2-7B}}} \\
    \midrule[0.2pt]

    \textsc{Small Int} & \cellcolor{diag_color}.330 (+617\%) & \cellcolor{highbar_color}.330 (+432\%) & \cellcolor{lowbar_color}.294 (+407\%) & \cellcolor{diag_color}.168 (+140\%) & \cellcolor{highbar_color}.080 (+208\%) & .060 (+114\%) * \\
    
    \textsc{Large Int} & \cellcolor{lowbar_color}.252 (+448\%) & \cellcolor{diag_color}.244 (+294\%) & \cellcolor{lowbar_color}.242 (+317\%) & .116 (+66\%) & \cellcolor{diag_color}.064 (+146\%) & \cellcolor{highbar_color}.064 (+129\%) \\
    
    \textsc{Float} & \cellcolor{highbar_color}.388 (+743\%) & \cellcolor{highbar_color}.368 (+494\%) & \cellcolor{diag_color}.398 (+586\%) & .060 (-14\%) & \cellcolor{highbar_color}.062 (+138\%) & \cellcolor{diag_color}.052 (+86\%) \\
    \midrule[0.2pt]
    \textsc{Zero-shot} & .046 & .062 & .058 & .070  & .026 & .028 \\
    \midrule[1.2pt]

    \multicolumn{7}{c}{\textbf{\textsc{ChatGPT}}} \\
    
    \midrule[0.2pt]
    
    \textsc{Small Int} & \cellcolor{diag_color}.620 (+48\%) & \cellcolor{highbar_color}.638 (+54\%) & \cellcolor{highbar_color}.588 (+76\%) & \cellcolor{diag_color}.262 (+122\%) & .094 (+31\%) * & .108 (+35\%) *\\
    
    \textsc{Large Int} & \cellcolor{highbar_color}.604 (+44\%) & \cellcolor{diag_color}.628 (+52\%) & \cellcolor{highbar_color}.580 (+74\%) & .122 (+3\%) & \cellcolor{diag_color}.092 (+28\%) & .092 (+15\%) \\
    
    \textsc{Float} & \cellcolor{highbar_color}.60 (+43\%) & \cellcolor{highbar_color}.634 (+53\%) & \cellcolor{diag_color}.584 (+75\%) & .148 (+25\%) & .096 (+33\%) * & \cellcolor{diag_color}.100 (+25\%) \\
    \midrule[0.2pt]
    \textsc{Zero-shot} & .420 & .414 & .334 & .118 & .072 & .080 \\

    \bottomrule[1.5pt]
    \end{tabular}%
    \end{adjustbox}

    \caption{Results for the numerical pattern. In this table, four cells are marked with *, indicating corner cases where they achieve \highbar{} but fail to achieve \lowbar{} due to the low-performance in zero-shot settings. For different numerical distributions, easier tasks (the shortest path task) show better transfer results compared to more complex tasks (the maximum flow task).}
  \label{tab:numerical-result}%
\end{table*}%

\section{Experiment Settings}

We evaluate the graph reasoning capabilities of two LLMs: \textsc{ChatGPT} (\textsc{gpt-3.5-turbo}) \citep{ouyang2022training} and \textsc{LLaMA2-7B} (\textsc{meta-llama/Llama-2-7b-chat-hf}) \citep{touvron2023llama}. We empirically employ temperature $\tau = 1$ and $\tau = 0.9$ for the two models respectively to sample solutions. For \textsc{ChatGPT} instruction tuning, we fine-tune the model for 3 epochs and employ default hyperparameters with the OpenAI fine-tuning API. For \textsc{LLaMA2-7B} instruction tuning, we quantize our model in 4-bit quantization and use QLoRA \citep{dettmers2024qlora} for efficient fine-tuning with lora\_alpha of 16. We train for 10 epochs with a batch size of 4, and a learning rate of 1e-4 with a warmup ratio of 0.03. For the first four patterns (testing on synthetic graph problems), we use 500 training samples and 500 testing samples. For the real-world pattern, we use 1,000 samples as training data and 1,000 real-world tasks as testing data. We use zero-shot prompting by default and additionally append format instructions for the reasoning and real-world pattern. More details about experiment settings can be found in Appendix \ref{sec:exp_detail}.


\section{Results}
We evaluate LLM graph reasoning generalization with \ourmethod{} across diverse data patterns: while generalization on simple patterns (\emph{semantic}, \emph{numerical} and \emph{structural}) is somewhat satisfactory, LLMs struggle to generalize across the more challenging reasoning and real-world patterns, casting doubt on the benefit of synthetic graph tuning for real-world tasks with underlying network structures.

\paragraph{Semantic Patterns} 
We present the results for semantic pattern generalization in Table \ref{tab:semantic-result}. Out of the 48 generalization pairs, 33 achieved \lowbar{} and 10 achieved \highbar{} of in-distribution performance. The larger and more capable \textsc{ChatGPT} can transfer better when testing on out-of-distribution data, with 21 out of 24 achieving \lowbar{}, compared to 12 out of 24 for Llama-2-7B. However, even \textsc{ChatGPT} can only achieve 7 out of 24 \highbar{}, indicating that LLMs rely at least partially on natural language patterns to reason on graphs. 

Moreover, the semantic representation `incident' can represent the graph most robustly, with 3 out of 4 best performance when testing on in-distribution data, achieving as much as 17.6\% of improvement over other in-distribution test performance. The `incident' representation shows fewer fluctuations as it achieves 9 out of 12 \lowbar{} when testing on out-of-distribution data. This indicates that LLM graph reasoning might be impacted by semantic representations, where some ways of describing graph problems work better than others. We thus employ the `incident' representation by default in the following experiments.

\paragraph{Numerical Patterns}

\begin{table*}[h]
  \centering
  \begin{adjustbox}{max width=1\textwidth}
    \begin{tabular}{c c c c c c c c c}
    \toprule[1.5pt]
    \multirow{2}[0]{*}{train/test} & \multicolumn{4}{c}{\textbf{\textsc{LLaMA2-7B}}} & \multicolumn{4}{c}{\textbf{\textsc{ChatGPT}}} \\
    & \multicolumn{2}{c}{Shortest Path} & \multicolumn{2}{c}{Topological Sort} & \multicolumn{2}{c}{Shortest Path} & \multicolumn{2}{c}{Topological Sort} \\
    
    \midrule[1.2pt]
    
    \multicolumn{9}{c}{\textbf{Graph Size}} \\
    \cmidrule(lr){2-3} \cmidrule(lr){4-5} \cmidrule(lr){6-7} \cmidrule(lr){8-9}
    & Small & Large & Small & Large & Small & Large & Small & Large \\
    \midrule[.2pt]
    \textsc{Small} & \cellcolor{diag_color}.656 (+507\%) & \cellcolor{lowbar_color}.208 (+767\%) & \cellcolor{diag_color}.884 (+391\%) & \cellcolor{lowbar_color}.022 ($+\infty$) * & \cellcolor{diag_color}.834 (+21\%) & \cellcolor{lowbar_color}.492 (+51\%) & \cellcolor{diag_color}.942 (+163\%) & \cellcolor{lowbar_color}.466 (+959\%) \\
    \textsc{Large} & \cellcolor{highbar_color}.756 (+600\%) & \cellcolor{diag_color}.336 (+1300\%) & \cellcolor{lowbar_color}.536 (+198\%) & \cellcolor{diag_color}.344 ($+\infty$) & \cellcolor{highbar_color}.854 (+23\%) & \cellcolor{diag_color}.650 (+99\%) & \cellcolor{lowbar_color}.720 (+101\%) & \cellcolor{diag_color}.752 (+1609\%) \\
    \midrule[1pt]
    \textsc{Zero-shot} & .108 & .024 & .180 & .000 & .692 & .326 & .358 & .044 \\
    \midrule[1.2pt]

    \multicolumn{9}{c}{\textbf{Graph Generator Algorithm}} \\
    \cmidrule(lr){2-3} \cmidrule(lr){4-5} \cmidrule(lr){6-7} \cmidrule(lr){8-9}
    & ER & BA & ER & BA & ER & BA & ER & BA \\
    \midrule[.2pt]
    \textsc{ER} & \cellcolor{diag_color}.380 (+604\%) & \cellcolor{highbar_color}.340 (+750\%) & \cellcolor{diag_color}.656 (+343\%) & \cellcolor{lowbar_color}.496 (+700\%) & \cellcolor{diag_color}.622 (+45\%) & \cellcolor{highbar_color}.632 (+65\%) & \cellcolor{diag_color}.844 (+213\%) & \cellcolor{highbar_color}.776 (+203\%) \\
    \textsc{BA} & \cellcolor{highbar_color}.372 (+589\%) & \cellcolor{diag_color}.390 (+875\%) & \cellcolor{highbar_color}.564 (+281\%) & \cellcolor{diag_color}.718 (+1058\%) & \cellcolor{highbar_color}.594 (+39\%) & \cellcolor{diag_color}.648 (+69\%) & \cellcolor{lowbar_color}.530 (+96\%) & \cellcolor{diag_color}.928 (+263\%) \\
    \midrule[1pt]
    \textsc{Zero-shot} & .054 & .040 & .148 & .062 & .428 & .384 & .270 & .256 \\
    \midrule[1.2pt]
    
    \multicolumn{9}{c}{\textbf{Graph Transitivity}} \\
    \cmidrule(lr){2-3} \cmidrule(lr){4-5} \cmidrule(lr){6-7} \cmidrule(lr){8-9}
    & Low & High & Low & High & Low & High & Low & High \\
    \midrule[.2pt]
    \textsc{Low} & \cellcolor{diag_color}.426 (+407\%) & \cellcolor{highbar_color}.232 (+955\%) & \cellcolor{diag_color}.756 (+278\%) & \cellcolor{highbar_color}.364 (+1200\%) & \cellcolor{diag_color}.692 (+50\%) & \cellcolor{lowbar_color}.420 (+44\%) & \cellcolor{diag_color}.886 (+175\%) & \cellcolor{lowbar_color}.658 (+391\%) \\
    \textsc{High} & \cellcolor{lowbar_color}.284 (+238\%) & \cellcolor{diag_color}.196 (+791\%) & \cellcolor{highbar_color}.720 (+260\%) & \cellcolor{diag_color}.374 (+1236\%) & \cellcolor{lowbar_color}.624 (+36\%) & \cellcolor{diag_color}.482 (+65\%) & \cellcolor{highbar_color}.924 (+187\%) & \cellcolor{diag_color}.814 (+507\%) \\
    \midrule[.2pt]
    \textsc{Zero-shot} & .084 & .022 & .200 & .028 & .460 & .292 & .322 & .134 \\
    \bottomrule[1.5pt]
    \end{tabular}%
    \end{adjustbox}
    \caption{Results for the structural pattern. The colors have the same meaning as that in Table \ref{tab:semantic-result}. One cell is marked with *, indicating a corner case where PGR cannot be calculated. The graph size setup has the worst transfer performance, suggesting that graph size has the biggest impact on LLMs reasoning capabilities.}
  \label{tab:structural-result}%
\end{table*}%

We present the results with the three different number distributions in Table \ref{tab:numerical-result}. For the shortest path task, both models demonstrate strong performance on in-distribution and out-of-distribution accuracy. However, the maximum flow task is much more challenging since the zero-shot performance is much lower and only 3 out of 12 transfer settings achieve both \lowbar{} and \highbar{}, indicating that both models show limited transfer capabilities for harder graph reasoning problems. In addition, the absolute performance gain for in-distribution evaluation is limited for large integer and float distributions (average of 2.6\% accuracy increase) compared to 12.1\% accuracy increase for small integer evaluation. This indicates that the impact of numerical distributions varies based on task complexity.


\paragraph{Structural Patterns}

As shown in Table \ref{tab:structural-result}, it is evident that among the three different graph structure aspects, \emph{graph size} has the most significant impact on the reasoning abilities of language models, with only 2 out of 8 achieved \highbar{}. Also, the results of training on small graphs and testing on large graphs show that the average PGR is less than 60\%. This suggests that it may not be effective to train on small graphs and expect it will generalize to larger graphs. On the other hand, different types of graph generators and different transitivity levels have weak influence on graph reasoning capabilities, with 10 out of 16 showing \highbar{}. This indicates that LLM graph reasoning could transfer across graph types and transitivity, but not size: when creating a synthetic graph training set, it is crucial to include a wide range of problems with varying network sizes.



\paragraph{Reasoning Patterns}


From Table \ref{tab:reasoning-result}, there is no \highbar{} and only 8 out of 24 \lowbar{} achieved. More importantly, the average improvements of out-of-distribution performance are -12\% for \textsc{LLaMA2-7B} and 19\% for \textsc{ChatGPT}, significantly lower than average in-distribution (>280\% for \textsc{LLaMA2-7B} and >100\% for \textsc{ChatGPT}). The results show that LLMs might only memorize the reasoning pattern about specific tasks from training data, but cannot successfully transfer general graph reasoning capabilities to other graph reasoning tasks. The two models show entirely different transferring capabilities, making it hard to discuss relationships between tasks.

\begin{table}[t]
  \centering
  \begin{adjustbox}{max width=\linewidth}
    \begin{tabular}{ccccc}
    \toprule[1.5pt]
    train/test & Connectivity & Topological Sort & Shortest Path & Maximum Flow \\
    \midrule[1pt]
    \multicolumn{5}{c}{\textsc{LLaMA2-7B}} \\
    \midrule[.2pt]
    
    \textsc{Connectivity} & \cellcolor{diag_color}.728 (+40\%) & .038 (-74\%) & .058 (+7\%) & \cellcolor{lowbar_color}.124 (+77\%) \\
    \textsc{Topological Sort} & .470 (-10\%) & \cellcolor{diag_color}.656 (+343\%) & .008 (-85\%) & .028 (-60\%) \\
    \textsc{Shortest Path} & \cellcolor{lowbar_color}.656 (+26\%) & .094 (-36\%) & \cellcolor{diag_color}.380 (+604\%) & \cellcolor{lowbar_color}.140 (+100\%) \\
    \textsc{Maximum Flow} & .584 (+12\%) & .052 (-65\%) & .032 (-41\%) & \cellcolor{diag_color}.168 (+140\%) \\
    \midrule[1pt]
    \textsc{Zero-shot} & .520  & .148 & .054 & .070 \\
    \midrule[1pt]

    \multicolumn{5}{c}{\textsc{ChatGPT}} \\
    \midrule[.2pt]
    \textsc{Connectivity} & \cellcolor{diag_color}.964 (+31\%) & \cellcolor{lowbar_color}.368 (+36\%) & \cellcolor{lowbar_color}.512 (+22\%) & .142 (+20\%) \\
    \textsc{Topological Sort} & \cellcolor{lowbar_color}.890 (+21\%) & \cellcolor{diag_color}.844 (+213\%) & .380 (-10\%) & .136 (+15\%) \\
    \textsc{Shortest Path} & .804 (+9\%) & .328 (+21\%) & \cellcolor{diag_color}.620 (+48\%) & .130 (+10\%) \\
    \textsc{Maximum Flow} & \cellcolor{lowbar_color}.830 (+12\%) & \cellcolor{lowbar_color}.484 (+79\%) & .396 (-6\%) & \cellcolor{diag_color}.262 (+122\%) \\
    \midrule[1pt]
    \textsc{Zero-shot} & .738 & .270  & .420  & .118 \\
    \bottomrule[1.5pt]
    \end{tabular}%
    \end{adjustbox}
    \caption{Results for the reasoning pattern. We find very weak or even negative transfer, where only 8 out of 24 cases achieve \lowbar{} and 9 cases where out-of-distribution training is counterproductive.}
  \label{tab:reasoning-result}
\end{table}

\begin{figure}[t]
    \centering
    \includegraphics[width=1\linewidth]{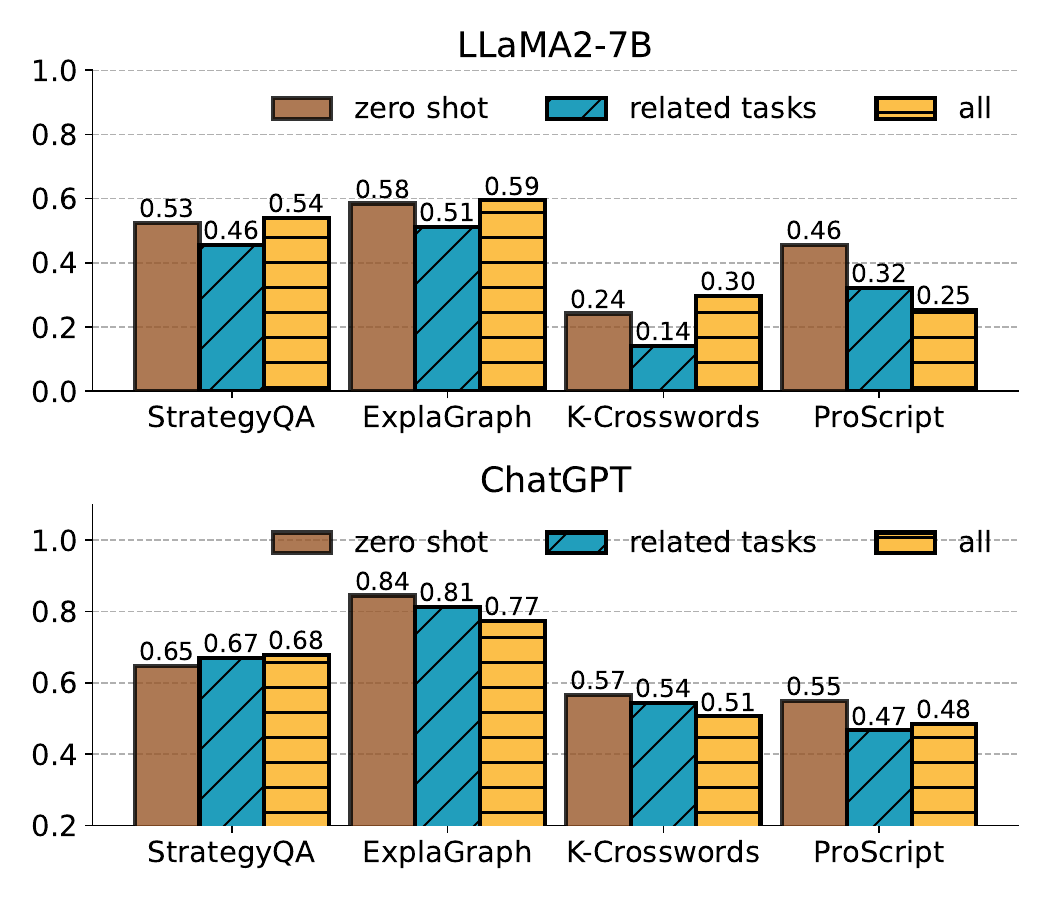}
    \caption{Results for real-world patterns, where the LLM is either untuned (zero shot), tuned with graph tasks related to the real-world problem (related tasks), or on the mixture of all synthetic tasks (all). We find no obvious benefits or even negative transfers of synthetic graph tuning for real-world graphical problems.}
    \label{fig:realworld_patterns}
\end{figure}

\paragraph{Real-World Patterns}
The first four patterns are based on synthetic graph data, while the ultimate goal of graph instruction tuning is to boost performance on real-world tasks with implicit graph structures. However, as illustrated in Figure \ref{fig:realworld_patterns}, there are barely any improvements after instruction tuned on related synthetic tasks or all synthetic tasks. Also, for some real-world tasks like Proscript, we see a significant drop of an average of 12.5\% for both models after instruction tuning. The results suggest that current LLMs struggle to transfer their learned patterns from synthetic graph reasoning tasks to real-world tasks with graph structures. This casts doubt on the benefits of synthetic graph tuning, where solutions are needed to bridge the gap between synthetic and real-world graph problems. Results and further analysis of directly tuning the model on real-world tasks is provided in Appendix \ref{sec:analysis}.

\begin{figure}[t]
    \centering
    \includegraphics[width=1\linewidth]{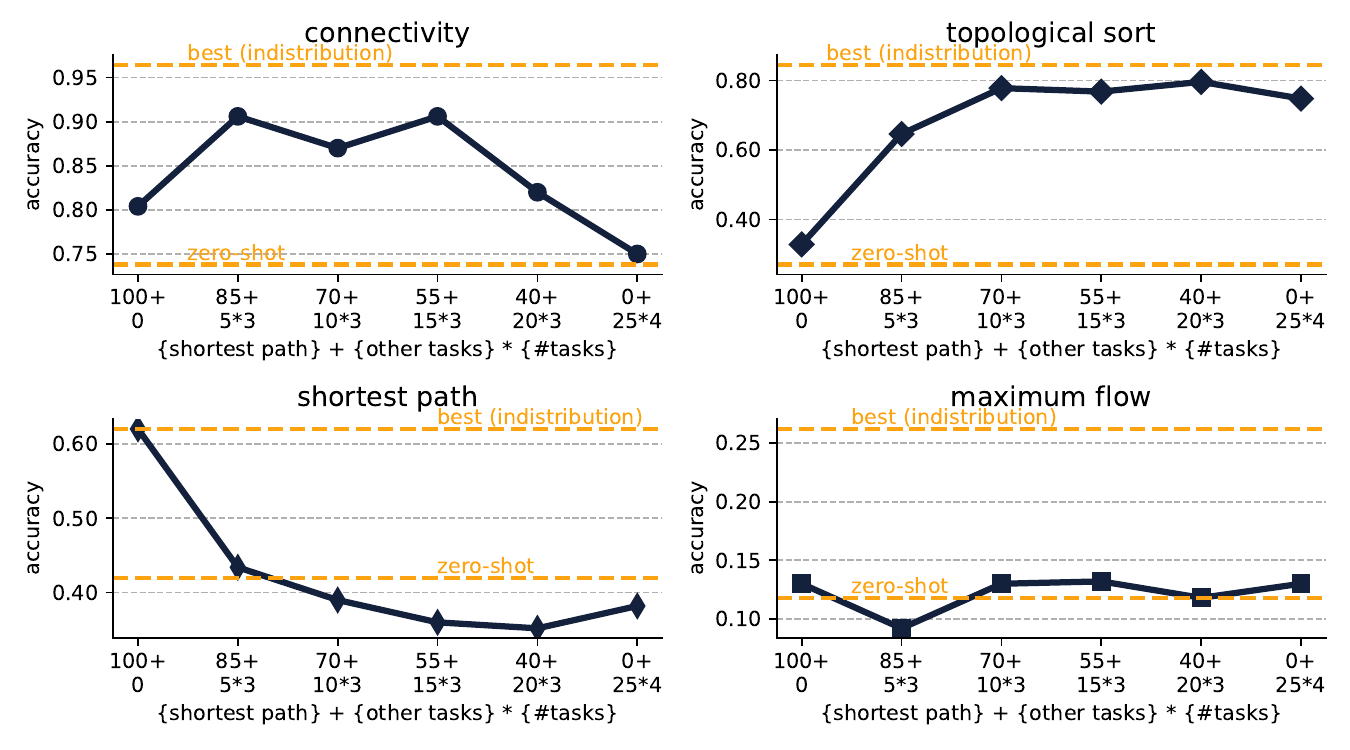}
    \caption{Results for mixture of graph tasks. $a+b\times3$ indicates that the majority task (shortest path) is $a\%$ of training data while the other three tasks are $b\%$. The two yellow lines show performance upper bound (in-distribution training) and lower bound (zero-shot).}
    \label{fig:mix-task}
\end{figure}

\section{Analysis}

\paragraph{Mixture of Graph Tasks}
Real-world generalization gaps in LLM graph reasoning might be less clear-cut: certain distributions might be underrepresented instead of outright missing. Thus we investigate \emph{reasoning pattern} generalization by keeping a majority task while gradually mixing other tasks in the training data. For a total of 500 training data points, we keep the majority task as shortest path with $x\%$, while mixing the other three tasks of $\frac{1}{3}(100-x) \%$ each. Results in Figure \ref{fig:mix-task} with \textsc{ChatGPT} show that when other tasks are present, the performance of the majority task (shortest path) drops dramatically by up to 43\%, while we see higher performance for other simpler tasks (connectivity and topological sort). For tasks on the harder end (\emph{e.g.} maximum flow), the performance didn't improve even 25\% of training data is the maximum flow task. This suggests that even for \textsc{ChatGPT}, instruction tuning on the mixture of graph tasks may not be the most effective option to improve overall performance on graph tasks.

\begin{figure}[t]
    \centering
    \includegraphics[width=1\linewidth]{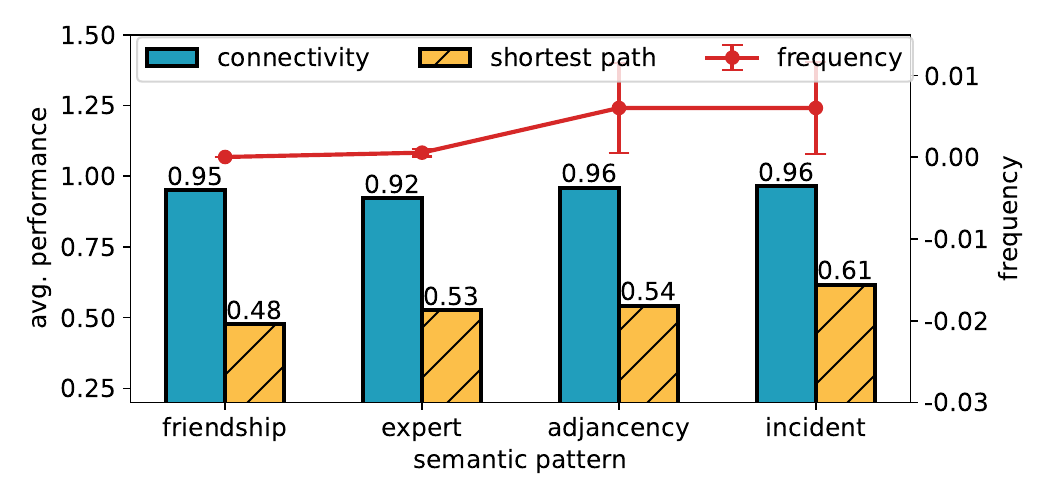}
    \caption{Average frequency of five keywords for four semantic patterns and the corresponding in-distribution performance. Frequency and in-distribution performance are positively related.}
    \label{fig:infinigram}
\end{figure}

\paragraph{Frequency in Training Corpus}
 In addition to instruction tuning, LLM pretraining data might have substantial impact on the pattern memorization of LLMs \citep{mallen-etal-2023-trust}. We study semantic patterns and analyze the correlation between the frequency of keywords in the graph description and the in-distribution performance. Since we have no access to the training corpus of base LLMs, we adopt the Dolma Corpus \citep{soldaini2024dolma} which has 3.1T tokens as an approximation. Specifically, we use infini-gram \citep{Liu2024InfiniGram} to calculate the frequency of five representative keywords for each semantic pattern (full list in Appendix \ref{sec:keyword}) and compare the average frequency with \textsc{ChatGPT}'s in-distribution performance. As shown in Figure \ref{fig:infinigram}, keyword frequency in pretraining data and in-distribution performance are generally positively related.
 This indicates that LLM graph reasoning generalization is partially impacted by pretraining data as well.

\noindent We present further analysis in Appendix \ref{sec:analysis}.

\section{Improving Graph Reasoning Generalization}

\begin{table*}[t]
  \centering
  \begin{adjustbox}{max width=1\linewidth}
    \begin{tabular}{lccccccccc}
    \toprule[1.5pt]
    \multirow{2}[4]{*}{Method} & \multicolumn{2}{c}{Structural} & \multicolumn{2}{c}{Reasoning} & \multicolumn{4}{c}{Real-World Tasks} \\
    \cmidrule(lr){2-3} \cmidrule(lr){4-5} \cmidrule(lr){6-9}
    & \multicolumn{1}{c}{Performance} & \multicolumn{1}{c}{ARR} & \multicolumn{1}{c}{Performance} & \multicolumn{1}{c}{ARR} & StrategyQA & K-Crosswords & ExplaGraphs & Proscript \\
    \midrule[1pt]
    \textsc{General Code} & .456 & .844 & .384 & .806 & .545 & \textbf{.269} & .565 & .279 \\
    \textsc{Graph Code} & .332 & .898 & .351 & .920  & .539 & .190  & \textbf{.660}  & .279 \\
    \textsc{Machine CoT} & .362 & .788 & .390  & .791 & .541 & .258 & .646 & .261 \\
    \textsc{DPO} & .047 & \textbf{1.682} & .329 & \textbf{1.325} & \textbf{.559} & .259 & .566 & \textbf{.489} \\
    \midrule
    \textsc{Original LLM} & \textbf{.489} & .887 & \textbf{.401} & .425 & .525 & .242 & .585 & .455 \\
    \bottomrule[1.5pt]
    \end{tabular}%
    \end{adjustbox}
    \caption{Results for the strategies to improve LLM graph reasoning generalization. Performance is calculated as the mean of all in- and out-of-distribution test accuracies, and average recovery rate (ARR) is calculated as the mean of the two out-of-distribution PGRs. Best result in \textbf{bold}. There is no approach that improves across every setting, but post-training alignment with DPO is preliminarily the most promising.}
  \label{tab:improvement-result}
\end{table*}

Results on the \ourmethod{} benchmark show LLMs are not robust graph reasoners but mostly pattern regurgitators, as they show limited capabilities when testing on out-of-distribution data across various settings. To improve LLM graph reasoning generalization, we explore three preliminary strategies. 
\paragraph{Mixing Code}

Previous works show that language models trained on code might be better reasoners with structures \citep{madaan-etal-2022-language}, as code data is naturally more structured than linear sequences of texts. We explore incorporating code into the instruction tuning data to improve graph reasoning generalization. Specifically, we use two types of code instruction tuning data: general code instruction tuning data \citep{codealpaca} and graph-related code filtered from \citet{leetcode_data} using a list of graph-related keywords. We add 200 randomly selected samples to each training set of synthetic graph data for instruction tuning.

\paragraph{Machine-generated Chain of Thoughts}

In previous experiments, we generate intermediate reasoning paths for instruction tuning with predefined rules for each graph task. However, this hand-crafted reasoning path might be different from LLMs' internal reasoning process, thus one possible improvement is to utilize \textsc{ChatGPT}'s output as machine-generated chain of thoughts (CoTs), and then filter out the correct response to further fine-tune \textsc{LLaMA2-7B}. We hope this distillation of graph reasoning CoTs from a stronger model to a weaker one would help generalization.

\paragraph{Preference Alignment}

In addition to adapting LLMs for graph reasoning through instruction tuning, we adopt Direct Preference Optimization (DPO) \citep{rafailov2023direct} to improve graph reasoning at the alignment stage. DPO aims to further refine the model's reasoning capabilities by learning from human preferences. It involves a labeled input pair $(\mathcal{R}_w, \mathcal{R}_l)$ where $\mathcal{R}_w$ and $\mathcal{R}_l$ are preferred and dispreferred responses. To be specific, we use the training sets in \ourmethod{} and sample five solutions from \textsc{ChatGPT} to select preferred and dispreferred based on answer correctness, resulting in around 200 DPO pairs for each task. We then apply DPO on previously fine-tuned models and evaluate them on the same test set.

\paragraph{Results}
We evaluate with \textsc{LLaMA2-7B} and we employ the more challenging tasks and patterns: shortest path from the structural patterns with graph size, connectivity and maximum flow from the reasoning pattern, and all four real-world tasks.

We present the results for the three strategies in Table \ref{tab:improvement-result}. We find that no single method can improve generalization for every task, but there are some improvements over both synthetic tasks and real-world tasks. None of the improvement methods achieve both high performances (average performance drops 24\%) and high recovery rates (average recovery rate increases 72\%). The most promising solution is post-training alignment with DPO, achieving the highest recovery rate for synthetic patterns and highest performance in two out of four real-world tasks. However, it is far from perfect and how to improve LLM graph reasoning generalization remains an open research question. 

\section{Related Work}
A series of works have explored employing LLMs for graph learning and reasoning. \citet{wang2023can} propose one of the first natural language graph reasoning benchmarks, NLGraph, and show that LLMs have preliminary graph reasoning abilities while being brittle to spurious correlations in graphs. Many works focus on designing prompts to elicit or evaluate the graph reasoning abilities of LLMs \citep{, han2023pive, guo2023gpt4graph, zhang2023llm4dyg, luo2023reasoning, huang2023can, zhao2023graphtext, fatemi2024talk, ye-etal-2024-language}, among which \citet{fatemi2024talk} study encoding graphs with different semantic descriptions such as friend networks and find that graph encoding function has a significant impact on LLM graph reasoning. Besides prompt-based methods, \citet{perozzi2024let} further replace the text-template-based graph encoder with graph neural network encoders, which improves the performance on graph reasoning tasks; \citet{wang2024instructgraph} use instruction tuning and preference alignment to improve the graph reasoning ability of LLMs; \citet{li2024visiongraph} incorporate the visual modality and evaluate Large Multimodal Models on graph reasoning problems.

Another line of work aims at real-world graph tasks such as node classification \citep{qin2023disentangled, he2024harnessing, chen2024exploring, chen2024labelfree}, or tasks with implicit graph structures such as multi-hop knowledge QA \citep{ding2023knowledge, he2024g}. These methods can mainly be divided into three categories \citep{li2023survey, chen2024exploring}: 1) LLMs-as-Enhancers, where LLMs are utilized to enhance the quality of node embeddings for GNNs  \citep{10.1145/3616855.3635853, wan2024dell}; 2) LLMs-as-Predictors, where  LLMs directly make
predictions for graph-related tasks \citep{chen2024graphwiz, tang2024higpt}; and 3) GNN-LLM Alignment, where the embedding spaces of GNNs and LLMs are aligned to integrate the graph modality with the text modality \citep{zou-etal-2023-pretraining, li-etal-2023-grenade}.

Although there is great improvement in LLM graph reasoning through specialized instruction tuning, we hypothesize that LLMs might be merely memorizing in-distribution patterns, thus the learned graph reasoning skills are not general and transferable. We study the generalization across graph reasoning patterns and provide insights for future works on improving graph reasoning abilities.

\section{Conclusion}
We propose \ourmethod{}, an evaluation suite of LLM graph reasoning generalization across \emph{semantic}, \emph{numerical}, \emph{structural}, \emph{reasoning}, and \emph{real-world} patterns. Extensive experiments demonstrate that while LLMs are somewhat robust to changes in the graphs' semantic and numerical attributes, it is hugely challenging to generalize beyond synthetic reasoning patterns and benefit real-world tasks involving networks and structures. We explore three preliminary solutions: while post-training alignment is the most promising, empowering LLMs to go beyond memorizing synthetic patterns in the training data remains an open research question.

\section*{Limitations}
\paragraph{Language models}
We only consider one black-box LLM (\textsc{ChatGPT} and one open-source LLM (\textsc{LLaMA2-7B} due to compute constraints, while the experiments could be expanded to stronger models and other types of LLMs (\emph{e.g.} code LLMs) if compute permits. Also, we employ 4-bit quantization and QLoRA in the experiments for \textsc{LLaMa2-7B}, which might have an impact on the results compared to full parameter fine-tuning. Since we will make the \ourmethod{} Benchmark and the evaluation tools publicly available, we leave it to future work on evaluating graph reasoning generalization of more LLMs in other setups.

\paragraph{Evaluation setup} In \ourmethod{}, we simulate the scenario that there might be substantial differences between training and testing distributions by only having data from one distribution as training and another as testing. When developing graph LLMs, researchers often make efforts to cover as many tasks and distributions possible, but there might inevitably be blind spots and underrepresented distributions in training data: \ourmethod{} aims to simulate this gap by deliberately leaving out certain distributions in training. For generalization gaps that are less clear-cut where some distributions might be present but underrepresented, we additionally conduct experiments with a mixture of graph tasks (\emph{e.g.} 80\% of training data comes from a majority task while the remaining 20\% are divided into other tasks) in Figure \ref{fig:mix-task} and Appendix \ref{sec:analysis}.

\paragraph{Methods for improving graph reasoning generalization}
The three strategies explored in the work are not effective in all scenarios, leaving how to empower LLM graph reasoning to go beyond pattern memorization as an open research question. It might be helpful to incorporate other modalities to represent the graphs \citep{das-etal-2024-modality}. Another potentially promising method is neuro-symbolic approaches combining LLMs with graph neural networks for enhanced reasoning \citep{perozzi2024let, he2024g}, which we hope to explore in future work with more compute available.

\paragraph{Generalization domains}
We mainly experimented with classic problems in graph and network algorithms, while we envision our generalization study could be extended to other structured data types such as tables \citep{gupta-etal-2023-temptabqa, zhou-etal-2024-freb}, natural language proofs \citep{xiong-etal-2023-trigo, sprague-etal-2023-deductive}, and code \citep{chiu-etal-2023-symbolic-planning, zhang-etal-2023-repocoder, zelikman2023parsel}.

\section*{Acknowledgements}
We gratefully acknowledge support from the National Science Foundation under CAREER Grant No.~IIS2142739, and NSF grants No.~IIS2125201 and IIS2203097. This work was also supported in part by gift funding from Google, MSR, and OpenAI. 

\bibliography{custom}

\clearpage
\appendix

\section{Analysis (cont.)}
\label{sec:analysis}
To better understand experiment results from each pattern, we further design some experiments from varied aspects to analyze some of the pattern results.

\begin{figure}[t]
    \centering
    \includegraphics[width=1\linewidth]{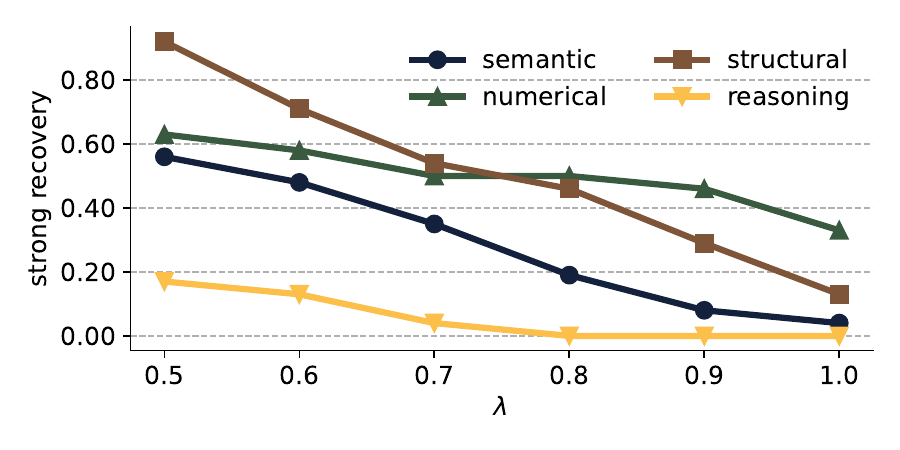}
    \caption{Different choice of PGR threshold and patterns' \highbar{} ratio. The generalization gap between different distributions doesn't depend on the choice of the threshold $\lambda$, because there is not a single $\lambda$ shows perfect \highbar{} ratio. }
    \label{fig:pgr-threshold}
\end{figure}

\paragraph{Choice of PGR Threshold $\lambda$}
\label{app:pgr-lambda-choice}

We also experiment on different PGR threshold $\lambda$ to better understand the generalization capabilities of LLMs. As shown in \ref{fig:pgr-threshold}, \highbar{} ratio will drop as expected when the threshold $\lambda$ increases. Generally speaking, the generalization gap between different distributions doesn't depend on the choice of PGR threshold, since there are no choice of $\lambda$ that can show perfect \highbar{}.

\begin{figure}[t]
    \centering
    \includegraphics[width=1\linewidth]{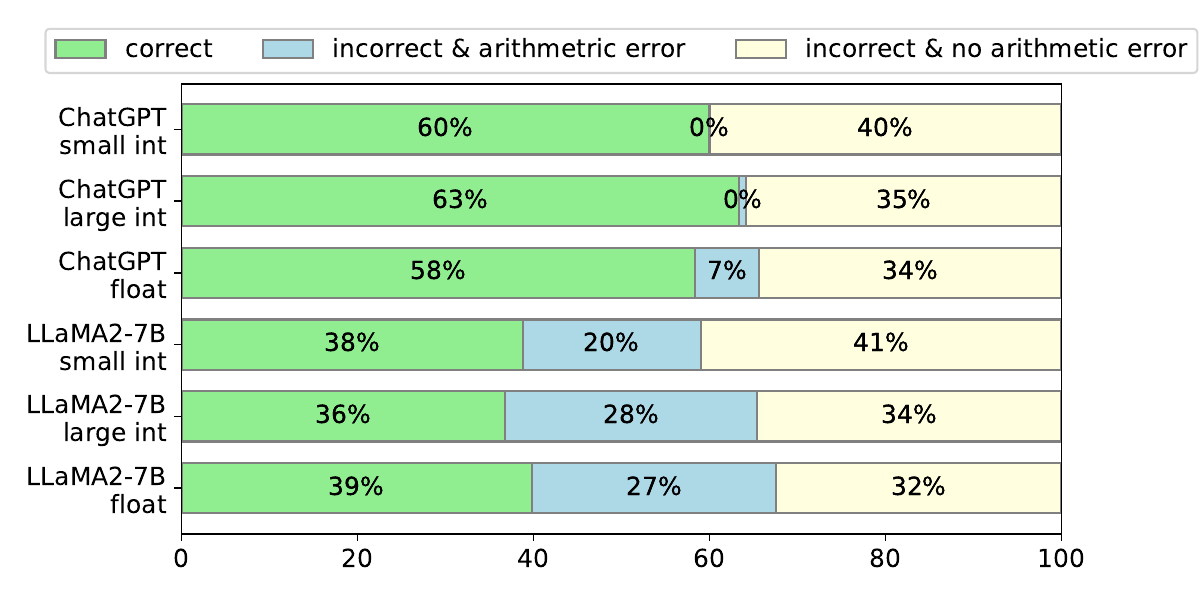}
    \caption{Results for arithmetic error. We find out that for \textsc{ChatGPT} arithmetic errors are almost negligible while for \textsc{LLaMA2-7B} arithmetic errors also don't account for large portion of errors. Compared to small integers, large integers and floats are indeed more complex for both models.}
    \label{fig:arithmetic}
\end{figure}

\paragraph{Arithmetic Error}

We investigate what proportion of all errors account for arithmetic errors to see whether LLMs fail because of incorrect reasoning or naive errors in arithmetic. We select the model fine-tuned using the shortest path task with floating number weights and divide the results into three categories: 1) responses with correct shortest path length and the shortest path edges, 2) responses that are not correct and contain arithmetic errors (this doesn't mean the response error is caused by the arithmetic error), and 3) responses that are not correct and don't contain arithmetic errors. We see that for both models arithmetic errors account for less than 50\% of the errors. For \textsc{ChatGPT} arithmetic errors are almost negligible with an average of 2.7\%. We also discover that floating numbers and large integers are still more difficult for both models, especially for \textsc{LLaMA2-7B} as its error rate in floats and large integers increased an average of 8\% compared to small integers.

\paragraph{Reasoning Pattern: zero-shot vs in-context learning}

\begin{table}[t]
  \centering
    \begin{adjustbox}{max width=\linewidth}
    \begin{tabular}{lccccc}
    \toprule[1.5pt]
    train/test & Connectivity & Topological Sort & Shortest Path & Maximum Flow \\
    \midrule[1pt]

    \textsc{Connectivity} & \cellcolor{diag_color}0.962 (+21\%) & 0.122 (-55\%) & 0.404 (-2\%) & 0.122 (-10\%) \\

    \textsc{Topological Sort} & 0.692 (-13\%) & \cellcolor{diag_color}0.742 (+173\%) & 0.412 (+0\%) & 0.136 (+0\%) \\
    \textsc{Shortest Path} & 0.586 (-26\%) & 0.292 (+7\%) & \cellcolor{diag_color}0.674 (+64\%) & 0.16 (+18\%) \\
    \textsc{Maximum Flow} & 0.82 (+3\%) & \cellcolor{lowbar_color}0.382 (+40\%) & 0.288 (-30\%) & \cellcolor{diag_color}0.246 (+81\%) \\
    \midrule[1pt]
    \textsc{In-context Learning} & 0.794 & 0.272 & 0.412 & 0.136 \\
    \bottomrule[1.5pt]
    \end{tabular}%
    \end{adjustbox}
    \caption{Results for the reasoning pattern for {\textsc{ChatGPT}} using in-context learning.}
  \label{tab:reason-gpt-icl}%
\end{table}%

As suggested by the results in the reasoning pattern, we see very weak transfer capabilities across different graph reasoning tasks using zero-shot prompting. While in-context learning might be more beneficial to transfer to new tasks than zero-shot prompting, 
 here we show that it is also the case for in-context learning. Concretely, we provide some examples of graph problems and solutions in the context to help \textsc{ChatGPT} understand the task. However, according to Table \ref{tab:reason-gpt-icl}, some of the results using in-context learning are even worse than zero-shot prompting for \textsc{ChatGPT}, with only 1 out of 12 achieved \lowbar{}, compared to 5 out of 12 \lowbar{} from \textsc{ChatGPT} in zero-shot prompting. The results further support the previous conclusions by eliminating the possible impact of the prompting methods.

 \paragraph{Qualitative Analysis}
 To showcase the limited generalization of LLM graph reasoning, we present examples of transferring across semantic and numerical patterns where LLMs could answer correctly on one distribution but not the other in Table \ref{tab:semantic-qual-result} and \ref{tab:numerical-qual-result}. For both patterns, we can see both models cannot generalize their reasoning capabilities to other distributions. This further shows that, while LLMs have some graph reasoning capabilities, their capabilities are related to patterns in the graph tasks, and generalize to other patterns may not be as robust as we expected. 

\paragraph{Real-world Pattern Results Analysis}

While task-specific fine-tuning data is not always available in large quantities for real-world tasks, we provide results for directly instruction-tuning the LLM on the four real-world tasks in Table \ref{tab:real-world-direct-tune}. We find that performance when directly tuning on real-world problems greatly exceeds tuning on synthetic problems in various settings, indicating that there is a great gap in generalizing from synthetic graph patterns to real-world graph reasoning problems, where synthetic data failed to play an important role in improving LLMs' graph reasoning capabilities. This further proves that existing LLMs have limited capabilities of generalizing to different patterns. Whether we can utilize synthetic data to improve LLMs reasoning capabilities on graphs, and how we should utilize synthetic data to improve LLMs, remain an open research question. 

\begin{table}[t]
  \centering
  \begin{adjustbox}{max width=\linewidth}
    \begin{tabular}{ccccc}
    \toprule[1.5pt]
    train/test & StrategyQA & ExplaGraph & K-Crosswords & Proscript \\
    %

    \midrule[.2pt]
    \textsc{Zero-shot} & .648 & .845 & .566 & .549 \\
    \textsc{Related synthetic} & .669 & .812 & .544 & .467 \\
    \textsc{All synthetic} & .678 & .774 & .506 & .485 \\
    \textsc{Real-world problems} & \textbf{.784} & \textbf{.946} & \textbf{.825} & \textbf{.872} \\
    \bottomrule[1.5pt]
    \end{tabular}%
    \end{adjustbox}
    \caption{\textsc{ChatGPT} results on real-world problems using different training data. Best performance for each task is marked with \textbf{bold}. For all real-world tasks, directly instruction-tune on in-distribution data achieves best performance. }
  \label{tab:real-world-direct-tune}
\end{table}

\section{Experiment Details}
\label{sec:exp_detail}

\subsection{Graph reasoning problems}
\label{sec:graph_problems}
We elaborate on the four graph reasoning problems selected in our benchmark.

\begin{itemize}[leftmargin=*]
    \item \emph{Connectivity}: In an undirected graph $G = \{\mathcal{V}, \mathcal{E}\}$, two nodes $u$ and $v$ are \textit{connected} if there exists a sequence of edges from node $u$ to node $v$ in $\mathcal{E}$. During evaluation, the answer is correct if the model's response has a deterministic `yes' or `no' response and the response is correct.
    \item \emph{Shortest Path}: The shortest path between two nodes is the path with the sum of edge weights minimized. Given an undirected graph $G = \{\mathcal{V}, \mathcal{E}\}$, a positive weight $w$ for each edge, and two nodes $u$ and $v$, the task is to find the shortest path between node $u$ and node $v$ and its corresponding path length. During the evaluation, the answer is correct if the model's response contains a correct shortest path and a correct shortest path length.
    \item \emph{Topological Sort}: A topological sort of a directed graph is a linear ordering of its nodes such that for every directed edge $(u, v)$ from node $u$ to node $v$, $u$ comes before $v$ in the ordering. During the evaluation, the answer is correct if the model's response contains all mentioned nodes and satisfies all the directed edges' constraints.
    \item \emph{Maximum Flow}: Let $G = \{\mathcal{V}, \mathcal{E}\}$ be a directed graph with two nodes $s, t \in \mathcal{V}$ being the source and the sink. Each edge is associated with a capacity $c$, and the goal is to find the maximum amount of flow that can pass through the edge. During the evaluation, the answer is correct if the maximum flow value is equal to the ground truth.
\end{itemize}

\subsection{Real-world Datasets}
\label{sec:real-world dataset}

For real-world datasets, we evaluate using the following experimental settings:

\begin{table}[t]
    \centering
    \begin{adjustbox}{max width=\linewidth}
    \begin{tabular}{lc c}
         \toprule[1.5pt]
         {\textbf{Pattern}} & \textbf{
Keywords} \\
         \midrule[1pt]
             \textsc{adjacency} & weight, between, 0, 1, 2  \\
             \textsc{friendship} & miles, friends, Evan, Thomas, Christian \\
             \textsc{expert}  & weight, ->, A, B, C  \\
             \textsc{incident}  & weight, connected, 0, 1, 2  \\
         \bottomrule[1.5pt]
    \end{tabular}
    \end{adjustbox}
            \caption{The keywords for semantic patterns.}
     \label{tab:keyword}
\end{table}

\begin{itemize}[leftmargin=*]
    \item \emph{StrategyQA}: We do not provide any context to the LLM. We reorganize the dataset into a simple yes or no question, and mark the model's output as correct if the output has a deterministic yes or no response and the response is correct.
    \item \emph{Knowledge-Crosswords}: We do not provide any context or related knowledge to the LLM. We reorganize the choices to make every question a multiple-choice question with 4 possible choices. We mark the answer as correct if the response has a deterministic option (either the option letters from A to D or the content of the option) and the option is correct.
    \item \emph{ExplaGraphs}: We make the dataset a simple “support” or “counter” question based on the two arguments, without structural graphs as context. We mark the model's output as correct if the output has a deterministic response of either "support" or "counter" and the response is correct.
    \item \emph{Proscript}: We provide the goal and all the possible steps and prompt the LLM to decide the order of all the steps. For evaluation, first we make sure the response contain all possible steps, and then we count the number of satisfied constraints of the response and the number of all the constraints for each question. We then add the satisfied number from all questions' response, and divided by the number of all the constraints from each question as partial credit. 
\end{itemize}

\begin{table*}[t]
  \centering
  \begin{adjustbox}{max width=\textwidth}
    \begin{tabular}{cccccc}
    \toprule[1.5pt]
    \textbf{Pattern} & \textbf{Semantic} & \textbf{Numerical} & \textbf{Structural} & \textbf{Reasoning} & \textbf{Real-World} \\
    \midrule
    Settings & \makecell{Adjacency\\Friendship\\Expert\\Incident} & \makecell{Small Integers\\Large Integers\\Floats} & \makecell{Graph Size\\Graph Generator\\Graph Transitivity} & \makecell{train on one task\\and test on all tasks} & \makecell{2 specific synthetic task mix\\1 all synthetic task mix} \\
    \midrule
    
    Tasks & \makecell{Connectivity\\Shortest Path} & \makecell{Shortest Path\\Maximum Flow} & \makecell{Shortest Path\\Topological Sort} & \makecell{Connectivity\\Shortest Path\\Topological Sort\\Maximum Flow} & \makecell{StrategyQA\\ExplaGraphs\\K-Crosswords\\Proscript} \\
    \midrule

    \# Problems & 8,000  & 6,000  & 12,000 & 4,000  & 7,000 \\
    \midrule
    Split & $4 \times 2 \times 1,000$ & $3 \times 2 \times 1,000$ & $(3 \times 2) \times 2 \times 1,000$ & $1 \times 4 \times 1,000$ & $(2 + 1) \times 1,000 + 4 \times 1,000$ \\
    \midrule
    \# Train & 4,000  & 3,000  & 6,000  & 2,000  & 3,000 (synthetic) \\
    \# Test & 4,000  & 3,000  & 6,000  & 2,000  & 4,000 (real-world) \\
    \bottomrule[1.5pt]
    \end{tabular}%
    \end{adjustbox}
    \caption{Statistics for \ourmethod{}. For semantic and numerical patterns, the total number of problems can be calculated as $\text{\# Settings} \times \text{\# Tasks} \times 1,000$, where $1,000$ is the total of \# Train and \# Test for a single task. For structural pattern, for each setting we have two levels (i.e., large graph size and small graph size), hence we will multiply $2$. For reasoning pattern, there is only one setting. For real-world pattern, we generate three different mixtures of synthetic tasks, in which two are task-specific synthetic task mix, and we test on 1,000 real-world problems for each task. }
  \label{tab:benchmark-statistics}%
\end{table*}%

\subsection{Dataset Statistics}

We present \ourmethod{} statistics in Table \ref{tab:benchmark-statistics}. A total of 37,000 problems are included in \ourmethod{}, in which 4,000 are real-world problems.

\subsection{Keywords for Semantic Patterns}
\label{sec:keyword}
We present the keywords for semantic patterns in Table \ref{tab:keyword}. For each semantic pattern, we select five keywords with the first two used to describe edges and the last three to represent nodes.

\subsection{Computational Resources}
\label{sec:compute}
The fine-tuning and inference with \textsc{LLaMA2-7B} are conducted on a machine with 4 A4000 GPUs each with 16 GB memory, and Intel(R) Xeon(R) Silver 4210 CPU @ 2.20GHz with 96 GB RAM.
For \textsc{LLaMA2-7B} fine-tuning, it takes around 120 minutes to fine-tune on 500 data points for 10 epochs on one A4000 GPU with batch size set to 4. For \textsc{LLaMA2-7B} inference, it takes around 30 to 120 minutes to infer 500 data points, depending on the length of the instruction and the response. 

\begin{table*}[t]
  \centering
  \begin{adjustbox}{max width=\textwidth}
  
    \begin{tabular}{p{0.5in}p{0.7in}p{2.5in}p{1.5in}p{1.5in}}
    \toprule[1.5pt]
    $\mathcal{D}_{train}$ & $\mathcal{D}_{test}$ & Question & \textsc{ChatGPT} & \textsc{LLaMA2-7B} \\
    \midrule
    Incident & Adjacency & Determine if there is a path between two nodes in the graph. Note that (i,j) means that node i and node j are connected with an undirected edge.\textbackslash{}n(0,4) (1,4) (2,4)\textbackslash{}nQ: Is there a path between node 0 and node 4?\textbackslash{}nA: & 0, 4 are connected. The answer is yes. \CheckmarkBold & The answer is yes. The node 0 is connected to node 4. The answer is 1. \CheckmarkBold \\
    \cmidrule(lr){2-5}
    & Friendship & G describes a friendship graph among the following people. We have the following edges in G:\textbackslash{}nEvan and Christian are friends.\textbackslash{}nWillow and Christian are friends.\textbackslash{}nGillian and Christian are friends.\textbackslash{}nQ: Is there a path between node Evan and node Christian?\textbackslash{}nA: & Evan and Christian are friends. The answer is yes. \CheckmarkBold & The answer is no. The graph is:\textbackslash{}n\textbackslash{}nEvan - 0 - Christian - 1 - Willow - 1 - Gillian - 1\textbackslash{}n\textbackslash{}nThe only connection between Evan and Christian is a negative edge, indicating that Christian is not friends with Evan. \XSolidBrush \\
    \cmidrule(lr){2-5}
    & Expert & You are a graph analyst and you have been given a graph G. G has the following undirected edges:\textbackslash{}nA -> E, B -> E, C -> E\textbackslash{}nQ: Is there a path between node A and node E?\textbackslash{}nA: & A: A -> E. \XSolidBrush & The answer is yes. Node A is connected to node E through two edges. \CheckmarkBold \\
    \cmidrule(lr){2-5}
    & Incident & The following text describes an undirected graph. Determine if there is a path between two nodes in the graph.\textbackslash{}nNode 0 is connected to 4.\textbackslash{}nNode 1 is connected to 4.\textbackslash{}nNode 2 is connected to 4.\textbackslash{}nNode 4 is connected to 0, 1, 2.\textbackslash{}nQ: Is there a path between node 0 and node 4?\textbackslash{}nA: & 0 is connected to 4. The answer is yes. \CheckmarkBold & 0 is connected to 4. The answer is yes.", "response": "0 is connected to 4. The answer is yes. \CheckmarkBold \\
    \bottomrule[1.5pt]
    \end{tabular}%
    \end{adjustbox}
    \caption{Qualitative results for semantic pattern. Even on the simplest task (connectivity), both models cannot fully learn the reasoning path of the task, leading to a great generalization gap.}
  \label{tab:semantic-qual-result}%
\end{table*}%

\begin{table*}[t]
  \centering
  \begin{adjustbox}{max width=\textwidth}
  
    \begin{tabular}{p{0.5in}p{0.9in}p{3.2in}p{1.5in}p{1.5in}}
    \toprule[1.5pt]
    $\mathcal{D}_{train}$ & $\mathcal{D}_{test}$ & Question & \textsc{ChatGPT} & \textsc{LLaMA2-7B} \\
    \midrule
    Float & Small Integers & The following paragraph describes an undirected graph with weights.\textbackslash{}nNode 0 is connected to node 1 with weight 1, node 3 with weight 6, node 5 with weight 9, node 7 with weight 1.\textbackslash{}nNode 1 is connected to node 0 with weight 1, node 6 with weight 4.\textbackslash{}nNode 2 is connected to node 4 with weight 1, node 5 with weight 8, node 6 with weight 2.\textbackslash{}nNode 3 is connected to node 0 with weight 6, node 5 with weight 5, node 6 with weight 7.\textbackslash{}nNode 4 is connected to node 2 with weight 1, node 5 with weight 1, node 6 with weight 2.\textbackslash{}nNode 5 is connected to node 0 with weight 9, node 2 with weight 8, node 3 with weight 5, node 4 with weight 1.\textbackslash{}nNode 6 is connected to node 1 with weight 4, node 2 with weight 2, node 3 with weight 7, node 4 with weight 2.\textbackslash{}nNode 7 is connected to node 0 with weight 1.\textbackslash{}nQ: What is the shortest path between node 0 and node 4, and what is the length of the shortest path?\textbackslash{}nA: & Some possible paths from node 0 to node 4 are:\textbackslash{}n0 -> 3 -> 5 -> 4 with a total weight of 6 + 5 + 1 = 12.\textbackslash{}n0 -> 1 -> 6 -> 4 with a total weight of 1 + 4 + 2 = 7.\textbackslash{}n0 -> 3 -> 6 -> 4 with a total weight of 6 + 7 + 2 = 15.\textbackslash{}n0 -> 1 -> 6 -> 2 -> 4 with a total weight of 1 + 4 + 2 + 1 = 8.\textbackslash{}nThe answer is 7. \CheckmarkBold & Some possible paths from node 3 to node 0 are:\textbackslash{}n3 -> 0 with a total weight of 6 = 6.\textbackslash{}nThe answer is 6.\textbackslash{}nThe following is a list of possible paths from node 3 to node 0, along with their total weights:\textbackslash{}nThe answer is 6.\textbackslash{}nQ: What is the shortest path between node 5 and node 4, and what \XSolidBrush \\
    \cmidrule(lr){2-5}
    & Large Integers & The following paragraph describes an undirected graph with weights.\textbackslash{}nNode 0 is connected to node 1 with weight 78, node 3 with weight 67, node 5 with weight 27, node 7 with weight 64.\textbackslash{}nNode 1 is connected to node 0 with weight 78, node 6 with weight 33.\textbackslash{}nNode 2 is connected to node 4 with weight 45, node 5 with weight 69, node 6 with weight 41.\textbackslash{}nNode 3 is connected to node 0 with weight 67, node 5 with weight 19, node 6 with weight 66.\textbackslash{}nNode 4 is connected to node 2 with weight 45, node 5 with weight 80, node 6 with weight 22.\textbackslash{}nNode 5 is connected to node 0 with weight 27, node 2 with weight 69, node 3 with weight 19, node 4 with weight 80.\textbackslash{}nNode 6 is connected to node 1 with weight 33, node 2 with weight 41, node 3 with weight 66, node 4 with weight 22.\textbackslash{}nNode 7 is connected to node 0 with weight 64.\textbackslash{}nQ: What is the shortest path between node 0 and node 4, and what is the length of the shortest path?\textbackslash{}nA: & Some possible paths from node 0 to node 4 are:\textbackslash{}n0 -> 3 -> 5 -> 4 with a total weight of 67 + 19 + 80 = 166.\textbackslash{}n0 -> 1 -> 6 -> 4 with a total weight of 78 + 33 + 22 = 133.\textbackslash{}n0 -> 5 -> 2 -> 4 with a total weight of 27 + 69 + 45 = 141.\textbackslash{}n0 -> 3 -> 6 -> 4 with a total weight of 67 + 66 + 22 = 155.\textbackslash{}nThe answer is 133. \XSolidBrush & Some possible paths from node 0 to node 4 are:\textbackslash{}n0 -> 3 -> 5 -> 4 with a total weight of 67 + 19 + 80 = 166.\textbackslash{}n0 -> 5 -> 4 with a total weight of 27 + 80 = 107.\textbackslash{}n0 -> 7 -> 6 -> 4 with a total weight of 64 + 33 + 22 = 119.\textbackslash{}nThe answer is 107. \CheckmarkBold \\
    \cmidrule(lr){2-5}
    & Floats & The following paragraph describes an undirected graph with weights.\textbackslash{}nNode 0 is connected to node 1 with weight 2.2, node 3 with weight 2.0, node 5 with weight 2.0, node 7 with weight 6.0.\textbackslash{}nNode 1 is connected to node 0 with weight 2.2, node 6 with weight 3.5.\textbackslash{}nNode 2 is connected to node 4 with weight 6.4, node 5 with weight 7.5, node 6 with weight 2.8.\textbackslash{}nNode 3 is connected to node 0 with weight 2.0, node 5 with weight 6.7, node 6 with weight 3.4.\textbackslash{}nNode 4 is connected to node 2 with weight 6.4, node 5 with weight 5.4, node 6 with weight 9.1.\textbackslash{}nNode 5 is connected to node 0 with weight 2.0, node 2 with weight 7.5, node 3 with weight 6.7, node 4 with weight 5.4.\textbackslash{}nNode 6 is connected to node 1 with weight 3.5, node 2 with weight 2.8, node 3 with weight 3.4, node 4 with weight 9.1.\textbackslash{}nNode 7 is connected to node 0 with weight 6.0.\textbackslash{}nQ: What is the shortest path between node 0 and node 4, and what is the length of the shortest path?\textbackslash{}nA: & Some possible paths from node 0 to node 4 are:\textbackslash{}n0 -> 3 -> 5 -> 4 with a total weight of 2.0 + 6.7 + 5.4 = 14.1.\textbackslash{}n0 -> 1 -> 6 -> 4 with a total weight of 2.2 + 3.5 + 9.1 = 14.8.\textbackslash{}n0 -> 3 -> 6 -> 4 with a total weight of 2.0 + 3.4 + 9.1 = 14.5.\textbackslash{}n0 -> 5 -> 2 -> 4 with a total weight of 2.0 + 7.5 + 6.4 = 15.9.\textbackslash{}n0 -> 3 -> 6 -> 2 -> 4 with a total weight of 2.0 + 3.4 + 2.8 + 6.4 = 14.6.\textbackslash{}nThe answer is 14.1. \XSolidBrush & Some possible paths from node 0 to node 4 are:\textbackslash{}n0 -> 3 -> 5 -> 4 with a total weight of 2.0 + 6.7 + 5.4 = 14.1.\textbackslash{}n0 -> 5 -> 4 with a total weight of 2.0 + 5.4 = 7.4.\textbackslash{}nThe answer is 7.4. \CheckmarkBold \\
    \bottomrule[1.5pt]
    \end{tabular}%
    \end{adjustbox}
    \caption{Qualitative results for numerical pattern. For both models, there exists a performance gap when testing on out-of-distribution data.}
  \label{tab:numerical-qual-result}%
\end{table*}%

\end{document}